%% file: ijcai22.tex
\documentclass{article}
\usepackage{ijcai22}

\usepackage{times}
\usepackage{soul}
\usepackage{color}
\usepackage{url}
\usepackage[hidelinks]{hyperref}
\usepackage[utf8]{inputenc}
\usepackage[small]{caption}
\usepackage{graphicx}
\usepackage{amsmath}
\usepackage{amsthm}
\usepackage{booktabs}
\usepackage{algorithm}
\usepackage{algorithmic}
\usepackage{multirow}
\usepackage{subfigure}
\usepackage{CJKutf8}
\usepackage{natbib}
\title{A Unified Strategy for Multilingual Grammatical Error Correction \\ with Pre-trained Cross-Lingual Language Model}

\author{
Xin Sun$^1$\thanks{This work was done during the author's internship at MSR Asia}
\and
Tao Ge$^2$\and
Shuming Ma$^2$\and
Jingjing Li$^3$\and
Furu Wei$^2$\and
Houfeng Wang$^1$
\affiliations
$^1$MOE Key Lab of Computational Linguistics, School of EECS, Peking University\\
$^2$Microsoft Research Asia\\
$^3$The Chinese University of Hong Kong
\emails
\{sunx5, wanghf\}@pku.edu.com \\
\{tage,shumma,fuwei\}@microsoft.com \\
lijj@cse.cuhk.edu.hk
}

\begin{document}

\maketitle

\begin{abstract}
Synthetic data construction of Grammatical Error Correction (GEC) for non-English languages relies heavily on human-designed and language-specific rules, which produce limited error-corrected patterns. In this paper, we propose a generic and language-independent strategy for multilingual GEC, which can train a GEC system effectively for a new non-English language with only two easy-to-access resources: 1) a pretrained cross-lingual language model (PXLM) and 2) parallel translation data between English and the language. 
Our approach creates diverse parallel GEC data without any language-specific operations by taking the non-autoregressive translation generated by PXLM and the gold translation as error-corrected sentence pairs. Then, we reuse PXLM to initialize the GEC model and pretrain it with the synthetic data generated by itself, which yields further improvement. We evaluate our approach on three public benchmarks of GEC in different languages. It achieves the state-of-the-art results on the NLPCC 2018 Task 2 dataset (Chinese) and obtains competitive performance on Falko-Merlin (German) and RULEC-GEC (Russian). Further analysis demonstrates that our data construction method is complementary to rule-based approaches.
\end{abstract}

\input{sections/introduction}
\input{sections/method}
\input{sections/experiment}
\input{sections/related}
\input{sections/conclusion}

\bibliographystyle{named}
\bibliography{ijcai22}
\input{sections/appendix}
\end{document}

%% file: sections/introduction.tex
\section{Introduction}

Grammatical Error Correction (GEC) is a monolingual text-to-text rewriting task where given a sentence containing grammatical errors, one needs to modify it to the corresponding error-free sentence. In recent years, pretraining on synthetic erroneous data and then fine-tuning on annotated sentence pairs has become a prevalent paradigm \citep{grundkiewicz2019minimally,lichtarge2019corpora,zhang2019sequence} in English GEC, advancing the state-of-the-art results with various novel data synthesis approaches~\citep{ge2018fluency,grundkiewicz2019minimally,kiyono2019empirical}. 


As GEC in other languages has drawn increasing attention ~\citep{flachs2021data,rothe2021simple}, synthetic erroneous data construction has been borrowed to non-English languages for improving the results given a lack of annotated data. For instance, the rule-based approaches obtain promising results \citep{grundkiewicz2019minimally,naplava2019grammatical,WANG_Chencheng:106}. However, these approaches require language-specific rules and confusion sets for word replacement based on expertise to simulate diverse linguistic phenomena across multiple languages, e.g., homophones in Chinese characters and morphological variants in Russian. Moreover, rule-based approaches always produce erroneous data with limited error-corrected patterns \citep{zhou2020improving}. 

\begin{figure}[t]
    \centering
    \includegraphics[width=\linewidth]{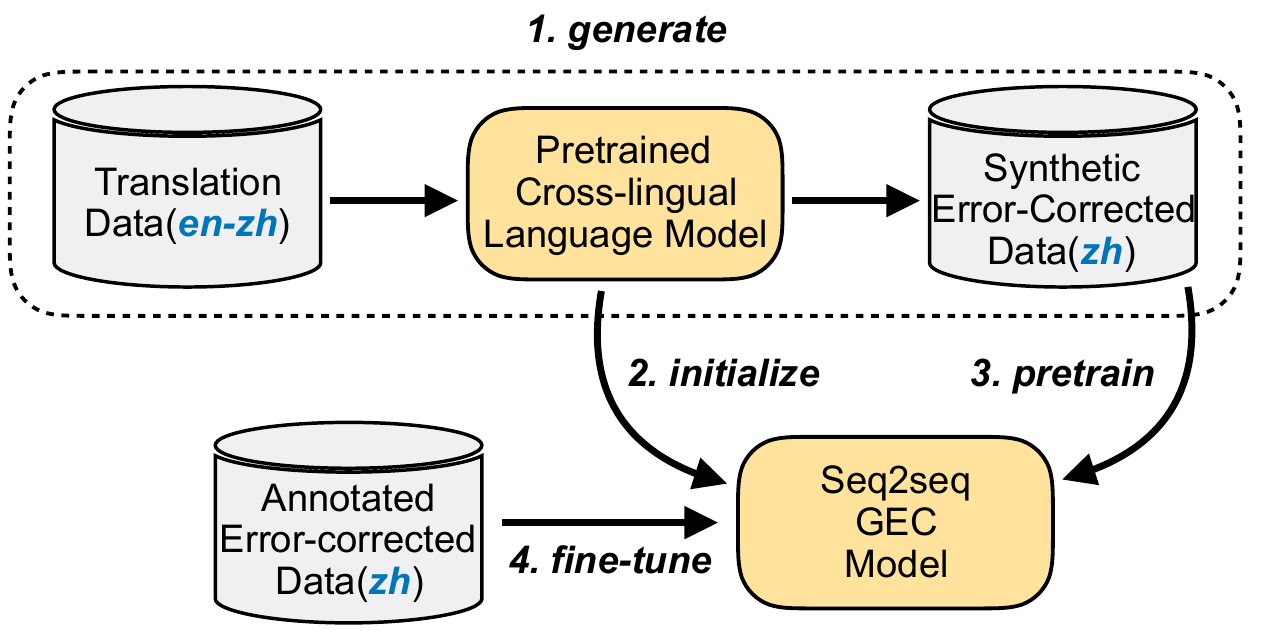}
    \caption{The overall framework of our approach. We use PXLM and a large-scale translation corpus to produce synthetic error-corrected sentence pairs. The seq2seq GEC model is initialized by PXLM and pretrained by the synthetic data. Then we fine-tune it with language-specific annotated GEC data. \textbf{En} and \textbf{Zh} denote English and Chinese, respectively.}
    \label{fig:intro}
\end{figure}

To address the above limitations, we propose a generic strategy for training GEC systems in non-English languages. Our approach is easily adapted to new languages if only provided with two relatively easy-to-obtain resources: 1) a pre-trained cross-lingual language model (PXLM); 2) the parallel translation data between English and the language. In this paper, we choose InfoXLM \citep{chi2020infoxlm} as the PXLM in our implementation.

Our approach consists of synthetic data construction and model initialization. Since InfoXLM was pre-trained with translation language modeling objective \citep{lample2019cross}, which requires the model to recover the masked tokens conditioned on the concatenation of a translation pair, it already possesses the capability of Non-Autoregressive Translation (NAT). That is, when presented with an English sentence, InfoXLM can provide a rough translation in dozens of non-English languages.

Compared with AT, NAT sacrifices translation quality due to the multimodality problem \citep{gu2017non,ghazvininejad2019mask}. When vanilla NAT performs independent predictions at every position, it tends to consider many possible translations of the sentence at the same time and output inconsistent results, such as token repetitions, missing or mismatch \citep{ran2020learning,du2021order}. Compared with pre-defined rules, such error-corrected patterns are more reasonable and diverse with large model capacity and dependency in sentence context. We regard the rough translation generated by InfoXLM as a source sentence and the gold translation as a corrected sentence for pretraining. To further improve the generalization ability of the seq2seq GEC model, we initialize the GEC model with InfoXLM and pretrain it with the synthetic data generated by itself. 


We conduct experiments on Chinese, German and Russian GEC benchmarks. Our approach achieves the state-of-the-art results on the NLPCC 2018 Task 2 dataset (Chinese) and obtains competitive performance on Falko-Merlin (German) and RULEC-GEC (Russian). The results also demonstrate that our approach can effectively complement rule-based corruption methods. 

The contributions of this paper are as follows:
\begin{itemize}
    \item We propose a unified strategy for GEC in the non-English languages consisting of synthetic data construction and model initialization.
    \item We propose a novel NAT-based synthetic data construction approach, which generates diverse error-corrected data for pretraining. To the best of our knowledge, it is the first to utilize the non-autoregressive translation ability of a PXLM for GEC erroneous data construction. The generated sentence pairs perform promising results alone and also nicely complement rule-based corruption methods.
    \item Our approach achieves the state-of-the-art performance on the Chinese benchmark and very competitive results for German and Russian benchmarks as well.
\end{itemize}

%% file: sections/method.tex
\section{Methodology}

\begin{figure*}[!ht]
    \centering
    \includegraphics[width=0.9\textwidth]{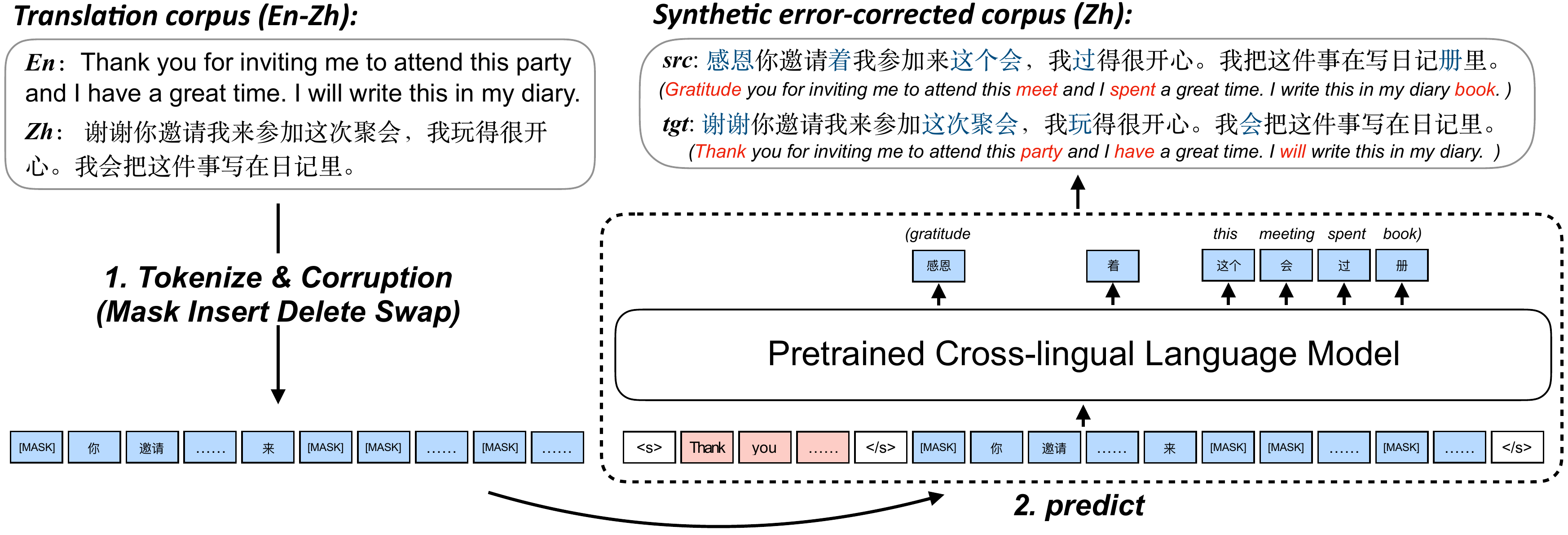}
    \caption{The overview of NAT-based data construction. Given a translation sentence pair (e.g., English-Chinese), our approach applies four operations (\textit{Mask}, \textit{Insert}, \textit{Delete} and \textit{Swap}) randomly to the non-English sentence. Then, PXLM predicts the possibility over the vocabulary at every masked position based on the concatenation of the English sentence and the corrupted sentence. Finally, we sample the predicted words and regard the recovered sequence as the source sentence containing grammatical errors and the gold non-English sentence as the corrected target sentence.}
    \label{fig:gen}
\end{figure*}

In this section, we present the unified strategy for non-English languages. At first, we briefly describe Translation Language Modeling (TLM) objective and Non-Autoregressive Translation (NAT) ability of InfoXLM. Then, we introduce two steps in our framework: NAT-based synthetic data construction and model initialization. Figure~\ref{fig:gen} shows the overview of our data construction approach.

\subsection{Background: Translation Language Modeling}

The basis of our data construction is the non-autoregressive translation ability of the InfoXLM, owing to its Translation Language Modeling (TLM) objective during pretraining. Given a sentence $x = x_1 \cdots x_{|x|}$ in the source language (e.g., English) and the corresponding translation $y = y_1 \cdots y_{|y|}$ in another language (e.g., Chinese), the input sequence of TLM is the concatenation of these two parallel sentences $S = \left<s\right>~x~\left</s\right>~y~\left</s\right>$ and some percentage of tokens are replaced with $\texttt{[MASK]}$ at random. Formally, let $M=\{m_1 , \cdots , m_{|M|}\}$ denote the positions of the masks:
\begin{gather}
    m_i \sim \mathrm{uniform}\{1, |x| + |y| + 3\} \quad \mbox{for} \; i=1 , \cdots , |M| \\
    S_M = \mathrm{replace}(S, M, \texttt{[MASK]})
\end{gather}
where the $\mathrm{replace}$ denotes the replacement operation at the certain positions. By leveraging bilingual context, the model is required to predict the original tokens with cross entropy loss.   The TLM loss is computed as:
\begin{equation}
    \mathcal{L}_{\text{TLM}} = -{\sum_{S\in \mathcal{T}}{\log{\prod_{ m \in M }{ p( S_m |S_{\setminus M} ) }}}}
\end{equation}
where $S_{\setminus M}=\{S_i\}_{i \not\in M}$ means tokens that are not included in the $M$ and $\mathcal{T}$ is the translation corpus.

To the extreme, we can use InfoXLM as a non-autoregressive translator. Specifically, we concatenate an English sentence $x$ with enough placeholders ($\texttt{[MASK]}$) as the input. InfoXLM is capable of translating it to other languages by predicting tokens at all masked positions in parallel. Formally, $M = \{|x| + 3, \cdots , |x| + |y| + 2\}$ denotes all target tokens are replaced with $\texttt{[MASK]}$ and the predicted translation $y^*$ is derived by maximizing the following equation:
\begin{align}
S' = \arg\max_{\boldsymbol{S_m}} \log{\prod_{ m \in M }{ p( S_m |S_{\setminus M} ) }} \\
y^* = \mathrm{replace}(S_M, M, S')
\end{align}
which infills the words with the highest probability.

In practice, following Mask-predict \citep{ghazvininejad2019mask}, we partially mask some percentage of target translation ($m \in \left[|x| + 3, |x| + |y| + 2\right]$) rather than all of them, which ensures the outputs are of appropriate quality.

\subsection{NAT-based Data Construction}
To generate diverse error-corrected sentences for GEC in a non-English language (e.g., Chinese), our approach utilizes sentence pairs of machine translation (e.g.,  English-Chinese parallel corpus). Our approach starts by adding noise to the target sentence and then masking sampled tokens. We feed the corrupted target sentence with the original English sentence as the input of InfoXLM. InfoXLM performs TLM predictions at every masked position. To obtain poor sentences containing grammatical errors, we randomly sample the word from the top predictions.  

Specifically, given a sentence $y$ in the target language, we select tokens for modification with a certain probability $p_{noise}$ and perform the following operations:

\textbf{Mask.} Replace the token with $\texttt{[MASK]}$ with a probability of $p_{mask}$.

\textbf{Insert.} Add a $\texttt{[MASK]}$ after the token with a probability of $p_{insert}$.

\textbf{Delete.} Delete the token with a probability of $p_{delete}$.

\textbf{Swap.} Replace the token with its right token with a probability of $p_{swap}$.

We get the noisy text $\widetilde{y} = \textsc{Noise}(y)$ and the corresponding positions of the masks $M$. Then, we concatenate the English sentence $x$ with the corrupted sequence $\widetilde{y}$ containing enough masked tokens as the input of InfoXLM. The predicted words are sampled for every $\texttt{[MASK]}$ according to the output distribution:
\begin{gather}
    y'_m \sim p(S_m |S_{\setminus M}) \quad \mbox{for} \; m \in \boldsymbol{M} \\
    y^* = \mathrm{replace}(\widetilde{y}, M, y')
\end{gather}
where we produce erroneous sentence by replacing $\texttt{[MASK]}$ with sampled tokens.

Our artificial corruption by four operations before TLM prediction improves the difficulty of translation. The independent assumption between target tokens brings in more errors and less fluency. The predicted words are sampled based on distribution rather than the best predictions to create more inconsistencies. It resembles the scenario where elementary language learners render a low-quality sentence when completing the cloze task. However, we only mask some percentage of target tokens and the English sentence restricts InfoXLM to recover original information, which ensures that the sampled tokens are plausible.

Since the recovered sentence contains diverse and reasonable word-level grammatical errors, we apply character-level corruption operations to add more spelling errors: 1) insert; 2) substitute; 3) delete; 4) swap-right; 5) change the casing. We call it \textbf{post edit}. Finally, we regard the gold translation as the corrected sentence and the corrupted prediction as the erroneous sentence. 

\subsection{Model Initialization}
To further improve the generalization ability of the GEC model, we use InfoXLM to initialize the seq2seq model. We follow \cite{ma2021deltalm} and use DeltaLM for multilingual GEC. DeltaLM is an InfoXLM-initialised encoder-decoder model trained in a self-supervised way. We continue pretraining it with synthetic data generated by our NAT-based approach.

Overall, our unified strategy exploits InfoXLM in two ways. We make use of its NAT ability to produce synthetic GEC data and its pretrained weights to initialize our GEC model. 

%% file: sections/experiment.tex
\section{Experiments}
\subsection{Data}
\begin{table}[]
    \centering
    \small
    \begin{tabular}{l|c|c|c|c}
    \textbf{Dataset} & \textbf{Language} & \textbf{Train} & \textbf{Valid} & \textbf{Test} \\
    \hline
    \textbf{NLPCC 2018 Task 2} & Chinese & 1.2M & 5000 & 2000 \\
    \textbf{Falko-Merlin} & German & 19237 & 2503 & 2337 \\
    \textbf{RULEC-GEC} & Russian & 4980 & 2500 & 5000 \\
    \hline
    \end{tabular}
    
    \caption{Statistics of the benchmarks for evaluation. The numbers in the table indicate the count of sentence pairs.}
    \label{dataset}
\end{table}
To validate the effectiveness of our approach in non-English languages, we conduct our experiments on three GEC datasets: NLPCC 2018 Task 2~\citep{zhao2018overview} in Chinese, Falko-Merlin~\citep{boyd2018using} in German and RULEC-GEC~\citep{rozovskaya2019grammar} in Russian. The statistics of the datasets are listed in Table~\ref{dataset}. We use the official Max-Match~\citep{dahlmeier2012better} scripts\footnote{\url{https://github.com/nusnlp/m2scorer}} to evaluate precision, recall and $F_{0.5}$.

\begin{table*}[t]
    \centering
    \begin{tabular}{l|cccc}
    \hline
    \multirow{2}{*}{\textbf{Model}} & \multicolumn{3}{c}{\textbf{NLPCC-18}} \\
    & $P$ & $R$ & $F_{0.5}$ \\
    \hline
    \bf YouDao \citep{fu2018youdao} & 35.24 & 18.64 & 29.91 \\
    \bf AliGM \citep{zhou2018chinese} & 41.00 & 13.75 & 29.36 \\
    \bf BLCU \citep{ren2018sequence} & \bf 47.23 & 12.56 & 30.57 \\
    \bf BERT-encoder \citep{wang2020chinese} & 41.94 & 22.02 & 35.51 \\
    \bf BERT-fuse \citep{wang2020chinese} & 32.20 & 23.16 & 29.87 \\
    \hline
    \bf Dropout-Src \citep{junczys2018approaching} & 39.08 & 18.80 & 32.15 \\
    \bf MaskGEC \citep{zhao2020maskgec} & 44.36 & 22.18 & 36.97 \\ 
    ~ - Our Implementation & 41.66 & 25.81 & 37.10 \\
    \hline
    \bf \cite{WANG_Chencheng:106} & 39.43 & 22.80 & 34.41 \\
    Rule(10M) & 44.66 & 26.54 & 39.30 \\
    Ours(10M) & 44.27 & 26.76 & 39.15  \\
    ~ - w/ DeltaLM & \bf 45.95 & \bf 27.94 & \bf 40.70 \\
    Ours(10M) + Confusion set & 45.17 & 26.11 & 39.42 \\
    Ours(5M) + Rule(5M) & 45.33 & \bf 27.61 & \bf 40.17 \\
    \hline
    \end{tabular}
    \caption{Performance of systems on the NLPCC-2018 Task 2 dataset. The results of different model architectures are shown at the top group. Different training strategies are shown in the middle. The approaches with pretraining are shown at the bottom. \textbf{Rule} denotes the synthetic data generated by rule-based corruption. \textbf{Ours} denotes data generated by our approach.}
    \label{nlpcc}
\end{table*}

For non-autoregressive translation generation, we use datasets of the WMT20 news translation task~\citep{barrault-etal-2020-findings} -- UN Parallel Corpus v1.0 \citep{ziemski2016united} for Chinese and Russian, the combination of Europarl v10\footnote{\url{http://www.statmt.org/europarl/v10/}}, ParaCrawl v5.1 \citep{banon-etal-2020-paracrawl} and Common Crawl corpus for German. We construct 10M synthetic sentence pairs in every language for pretraining and then fine-tune the GEC model on respective annotated datasets.

\subsection{Implementation Details}
Unless explicitly stated, we use Transformer (base) model in fairseq\footnote{\url{https://github.com/pytorch/fairseq}} as our GEC model. For Chinese, we construct a character-level vocabulary consisting of 7K tokens. We apply Byte Pair Encoding \citep{sennrich2016neural} to preprocess German and Russian sentences and obtain the vocabularies with size of 32K tokens, respectively. When using DeltaLM, we utilize its shared vocabulary of 250000 tokens based on the SentencePiece model~\citep{kudo2018sentencepiece}. During pretraining for German and Russian, following \cite{naplava2019grammatical}, we use source and target word dropouts and edit-weighted MLE \citep{junczys2018approaching}. We leave the detailed hyperparameters in the supplementary notes.

\begin{table*}[htb]
    \small
    \centering
    \begin{tabular}{l|c|ccc|ccc}
    \hline
    \multirow{2}{*}{\textbf{Model}} & 
    \textbf{Size} & \multicolumn{3}{c|}{\textbf{German}} & 
    \multicolumn{3}{c}{\textbf{Russian}} \\
    & \bf Layer, Hidden, FFN & $P$ & $R$ & $F_{0.5}$ & $P$ & $R$ & $F_{0.5}$ \\
    \hline
    \bf \cite{grundkiewicz2019minimally} & 6, 512, 2048 & 73.0 & 61.0 & 70.24 & 36.3 & \bf 28.7 & 34.46 \\
    \bf \cite{naplava2019grammatical} $\clubsuit$ & 6, 512, 2048 & \bf 78.11 & 59.13 & \bf 73.40 & 59.13 & 26.05 & 47.15 \\
    \bf \cite{rothe2021simple} $\spadesuit$ & 12, 768, 2048 & - & - & 69.21 & - & - & 26.24 \\
    Rule(10M) & 6, 512, 2048 & 73.71 & 59.28 & 70.29 & 49.38 & 23.49 & 40.46 \\
    Ours(10M) & 6, 512, 2048 & 73.86 & 60.74 & 70.8 & 57.96 & 23.51 & 44.82\\
    ~ - w/ DeltaLM & 12-6, 768, 2048 & \bf 75.59 & \bf 65.19 & \bf 73.25 & \bf 59.31 & 27.07 & \bf 47.90 \\
    Ours(5M) + Rule(5M) & 6, 512, 2048 & 74.31 & \bf 61.46 & 71.33 & \bf 61.40 & \bf 27.47 & \bf 49.24 \\
    \hline
    \bf \cite{flachs2021data} & 6, 1024, 4096 & - & - & 69.24 & - & - & 44.72 \\ 
    \bf \cite{naplava2019grammatical} $\clubsuit$ & 6, 1024, 4096 & \bf 78.21 & \bf 59.94 & \bf 73.71 & \bf 63.26 & \bf 27.50 & \bf 50.20 \\
    \bf \cite{katsumata2020stronger} & 12, 1024, 4096 & 73.97 & 53.98 & 68.86 & 53.50 & 26.35 & 44.36 \\
    \bf \cite{rothe2021simple} $\spadesuit$ & 24, 4096, 10240 & - & - & \bf 75.96 & - & - & \bf 51.62 \\
    
    \hline
    \end{tabular}
    \caption{Performance of systems on German and Russian datasets. \textbf{Layer}, \textbf{Hidden} and \textbf{FFN} denote depth, embedding size and feed forward network size of Transformer. \textbf{12-6} denotes that DeltaLM-initialized model has a 12-layer encoder and a 6-layer decoder. The top and bottom group shows the results of base-scale models and large-scale models, respectively. $\clubsuit$ Our re-implementation of this approach is \textbf{Rule(10M)}, whose results are inferior to \textbf{Ours(10M)}. We use synthetic data generated by their released codes and the same training strategy as ours. $\spadesuit$ It is good-performing with large-scale model size (up to 11B parameters) and our approach is better than its variant with similar model size.} 
    \label{de_ru}
\end{table*}

\subsection{Baselines}
Most of the previous studies for Chinese GEC focus on model architecture or training strategy, which are orthogonal with our synthetic data construction method. For example, \textbf{YouDao} \citep{fu2018youdao} combines five hybrid correction models and a language model together. \textbf{AliGM} \citep{zhou2018chinese} combines NMT-based, SMT-based and rule-based models together. \textbf{BLCU} \citep{ren2018sequence} uses multi-layer convolutional seq2seq model \citep{gehring2017convolutional}. \textbf{BERT-encoder} \citep{wang2020chinese} initializes the encoder of seq2seq model with BERT \citep{kenton2019bert}. \textbf{BERT-fuse} \citep{wang2020chinese} incorporates BERT for additional features. As for training strategy, \textbf{Dropout-Src} \citep{junczys2018approaching} sets the full embeddings of randomly selected source words to 0 during the training process. \textbf{MaskGEC} \citep{zhao2020maskgec} performs dynamic masking method by substituting the source word with a padding symbol or other word. 

The most comparable approach is \cite{WANG_Chencheng:106}, which constructs pretraining data using the rule-based corruption method. For our approach, we implement MaskGEC during the fine-tuning stage. To make a fair comparison, we also construct synthetic data with \textbf{rule-based corruption} in the same setting as baseline. It incorporates a character-level confusion set \footnote{\url{http://nlp.ee.ncu.edu.tw/resource/csc.html}} and uses \textit{pypinyin}\footnote{https://github.com/mozillazg/python-pinyin} to perform homophony replacement.

For German and Russian, the main data construction method is rule-based corruption. \cite{grundkiewicz2019minimally} and \cite{naplava2019grammatical} build confusion sets with edit distance, word embedding or spell-checker (e.g., Aspell dictionary\footnote{\url{http://aspell.net/}}). 
\cite{flachs2021data} uitlizes Unimorph \citep{kirov2018unimorph} which provides morphological variants of words for word replacement operations. They also incorporate WikiEdits \citep{lichtarge2019corpora} and Lang8 \citep{mizumoto2011mining} as additional training resources. \cite{rothe2021simple} only applies language-agnostic operations without any confusion set. They pretrain a unified seq2seq model for 101 languages and fine-tune for respective languages. \cite{katsumata2020stronger} proposes to directly use mBART \citep{liu2020multilingual} without pretraining.


\subsection{Main Results}
Table~\ref{nlpcc} shows the performance of our approach and previous methods on the NLPCC-2018 Chinese benchmark. Our NAT-based synthetic data construction approach is comparable with the rule-based corruption approach.
We assume that 0.15 $F_{0.5}$ descend comes from that rule-based corruption leverages many useful confusion sets. When combined with one of the character-level confusion sets, our approach obtains 39.42 $F_{0.5}$ which outperforms the rule-based counterpart. If combining two data sources from the rule-based and NAT-based approaches, we obtain better performance which demonstrates two methods complement each other. Initializing the GEC model with DeltaLM achieves 40.70 $F_{0.5}$, which is the state-of-the-art result of the dataset. 

Table~\ref{de_ru} shows the performance for German and Russian datasets. In the same setting, our NAT-based synthetic approach outperforms rule-based corruption methods and most baselines with \textbf{two exceptions}. For instance, \cite{naplava2019grammatical} leverages more training strategies during fine-tuning phase such as mixing pretraining data and oversampled fine-tuning data, checkpoint averaging and so on. \cite{rothe2021simple} obtains the best performance with large-scale model capacity (up to 11B parameters) and more sentence pairs for pretraining. Our approach significantly outperforms its base-scale variant. Overall, the performance on the German and Russian datasets demonstrates the effectiveness of our unified strategy and NAT-based synthetic approach, which performs competitive results alone and also nicely complements rule-based corruption methods.

\begin{table}[htb]
    \centering
    \small
    \begin{tabular}{l|c}
    \hline
    \bf Method & $F_{0.5}$ \\
    \hline
    \bf Rule(Unimorph)$^\star$ & 60.87 \\
    \bf Rule(Aspell)$^\star$ & 63.49 \\
    \bf Rule(Combine)$^\star$ & 62.55 \\
    \hline
    \bf WikiEdits$^\star$ & 58.00 \\
    \bf Rule + WikiEdits$^\star$ & \bf 66.66 \\
    \hline
    \bf Back-translate & 61.37 \\
    \bf Round-trip translation & 62.91 \\
    \hline
    \bf Ours & \bf 69.17 \\
    \hline
    \end{tabular}
    \caption{$F_{0.5}$ scores of different data construction approaches on the German dataset. For the approaches with $\star$, their results are from \cite{flachs2021data}.}
    \label{aug_compare}
\end{table}

To make fair comparison with multiple synthetic construction approaches, we follow the experimental setting and model hyperparameters\footnote{We use the ``transformer$\_$clean$\_$big$\_$tpu'' setting.} in \cite{flachs2021data}. The results on the German dataset are shown in Table~\ref{aug_compare}. Our approach significantly outperforms commonly used synthetic methods such as the rule-based approach with Unimorph, Aspell word replacement and Wikipedia edits extracted from the revision history.
Although back-translate is effective for English \citep{xie2018noising,kiyono2019empirical}, it performs poorly with limited annotated sentence pairs to learn diverse error-corrected patterns. 
Round-trip translation utilizes the same translation corpus as us but achieves inferior performance since it usually produces sentences without grammatical errors.

\subsection{Ablation Study}

\begin{table}[htb]
    \centering
    \small
    \begin{tabular}{l|ccc}
    \hline
     \bf Method & $P$ & $R$ & $F_{0.5}$ \\
     \hline
    \bf Ours & \bf 73.86 & 60.74 & \bf 70.8 \\
    ~ - [MASK] replacement & 71.17 & 55.07 &  67.24 \\
    ~ - [MASK] insert & 72.52 & 59.00 & 69.34 \\
    ~ - post edit & 73.00 & \bf 61.36 & 70.34 \\
    ~ - bilingual constraint & 71.17 & 55.89 & 67.48 \\
    ~ w/ an autoregressive translator & 66.99 & 55.44 & 64.31 \\
    \hline
    \end{tabular}
    \caption{Performance of our approach with different schemes on the German dataset. - denotes removing the component or replacing it with the rule-based operation.}
    \label{ablation}
\end{table}

We further conduct an ablation study as shown in Table~\ref{ablation}.
Overall, we find that all of these variants perform worse than the original strategy. From the last row, PXLM is much better than a regular translation model under the same setup (i.e., training data and sample strategy). Our approach can control the degree of overlap and error while AT generates clean sentences without grammatical errors or have minimal overlap with the original sentence. The removal of NAT-based replacement and bilingual constraint also results in a significant degradation, which indicates substitution with similar semantic meanings plays a crucial role in our strategy.

\subsection{Error-type Analysis}

\begin{table}[]
    \centering
    \scalebox{0.85}{
    \begin{tabular}{l|ccccc}
    \hline
     \bf Method & & None & Rule & Ours & Both \\
     \hline
     \bf Error Type & Ratio & $F_{0.5}$ & $F_{0.5}$ & $F_{0.5}$ & $F_{0.5}$ \\
     \hline
     Punctuation & 14.9 & 58.67 & 72.41 & \bf 73.98 & 73.75\\
     Spelling & 14.0 & 43.89 & 76.73 & 77.64 & \bf 78.76 \\
     Other & 9.8 & 8.64 & 29.18 & \bf 35.66 & 31.51 \\
     Determiner:FORM & 9.7 & 58.43 & 80.62 & 81.39 & \bf 81.9 \\
     Orthography & 8.3 & 66.38 & 76.33 & 73.7 & \bf 77.72 \\
     Adposition & 5.6 & 28.15 & 52.3 & \bf 53.2 & 51.7 \\
     Determiner & 4.7 & 25 & 50.1 & 55.9 & \bf 57.1 \\
     Adjective:FORM & 4 & 57.14 & 81.44 & 82.29 & \bf 83.41 \\
     Pronoun & 3.9 & 19.44 & 47.9 & 45.8 & \bf 51.4 \\
    \hline
    \end{tabular}
    }
    \caption{Performance of synthetic data construction approaches on top 9 error types of the German dataset. }
    \label{error_res}
\end{table}

We analyze the GEC performance of data construction approaches on different error types. We use the German
extension \citep{boyd2018using} of the automatic error annotation tool ERRANT \citep{bryant2017automatic} for evaluation. Table~\ref{error_res} shows the $F_{0.5}$ score of top 9 error types on the German dataset. We can observe that our approach improves the model in all error types significantly compared with that trained from the scratch and outperforms that with rule-based corruption in 7 out of 9 error types. For example, the largest improvement comes from the `Other' type by 6.5 $F_{0.5}$ score, which is defined as the errors that do not fall into any other specific type, such as paraphrasing (e.g., \textit{feel happy} $\rightarrow$ \textit{be delighted}) \citep{bryant2017automatic}. Such error type is beyond pre-defined rules and hard to simulate \citep{zhou2020improving}. The second-most improvement of our approach comes from `Determiner' (e.g., \textit{few} $\rightarrow$ \textit{some}), since the NAT-based method can replace tokens with similar semantics based on bilingual context. This reason also applies to the improvement of `Adjective:FORM' (e.g., \textit{goodest} $\rightarrow$ \textit{best}). Besides, our approach is even good at `Adposition' (e.g., \textit{at} $\rightarrow$ \textit{in}), `Punctuation' and `Spelling', which are intuitively more suitable for rule-based methods.

Two exceptions are `Orthography' and `Pronoun'. `Orthography' refers to the error related with case or whitespace errors (e.g., \textit{Nexttime} $\rightarrow$ \textit{next time}), which the specific rules are able to simulate very well. `Pronoun' denotes the substitutes for the noun (e.g. \textit{you} $\rightarrow$ \textit{yourself}) which also fall into the strength of the rule-based approach with the help of a language-specific confusion set.  We also observe the combination of these two sources of synthetic data yields better results, which demonstrates that they are helpful to complement each other to enrich error-corrected patterns.

\begin{figure}
    \centering
    \subfigure[Rule-based corruption.] { \label{fig:corr}
    \includegraphics[width=0.4\columnwidth]{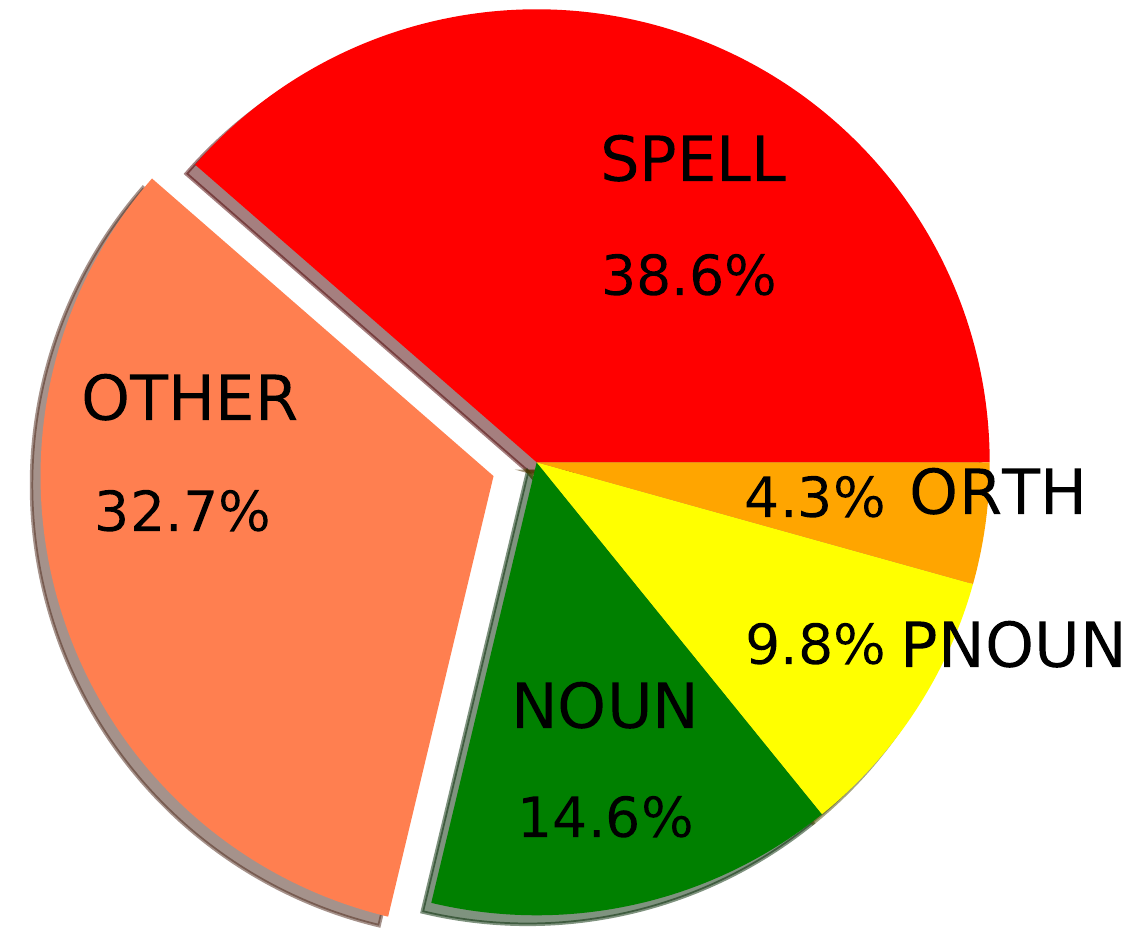}
}
    \subfigure[Our approach.] { \label{fig:nat}
    \includegraphics[width=0.35\columnwidth]{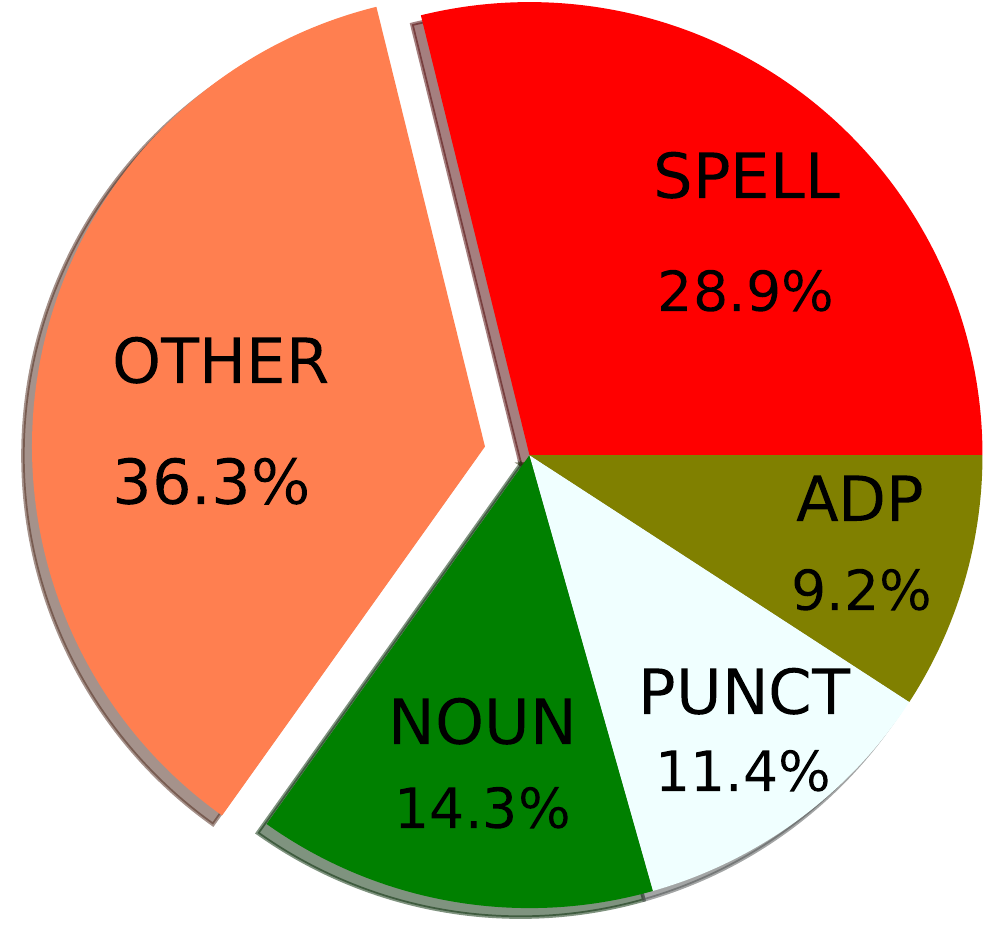}
}
    \caption{Top-5 error types distribution of different synthetic data construction approaches.}
    \label{err_dist}
\end{figure}

To verify our explanation, we present the top-5 error types distribution of rule-based corruption and our approach in Figure~\ref{err_dist}. Our approach yields more `Other' type errors compared with `Spell' errors, which may account for the significant improvement in that category. The large ratio of `Orthography' and `Pronoun' errors generated by rule-based corruption is consistent with its better performance on these two types. So are `Adposition' ($5$-th) and `Determiner' ($6$-th) generated by our approach.



\subsection{Case Study}

\begin{CJK*}{UTF8}{gbsn}
	\begin{table}[!ht]
		\centering
		\small
		\begin{tabular}{lp{6.7cm}}
		\hline
		\textbf{Source}  & 情况之所以特别严重，是因为这项法律在许多方面是其他法律的基础。 \\
		     & (The situation is particularly serious,  because in many ways this law is the basis of other laws.) \\
		  \textbf{BT} & 情况特别严重因为这项法律在许多\textcolor{blue}{其他}$_1$的方面是其他法律的基础。 \\
		  & (The situation is particularly serious because in many \textcolor{blue}{other}$_1$ ways this law is the basis of other laws.) \\
		  \textbf{Rule} & 情况之所以\textcolor{blue}{特别}$_1$，是因为这项法律\textcolor{blue}{在方面}$_2$是其他\textcolor{blue}{律法}$_3$的基础。\\
		   & (The situation is \textcolor{blue}{particular}$_1$, because \textcolor{blue}{in ways}$_2$ this law is the basis of other \textcolor{blue}{laws}$_3$.) \\
		  \textbf{RT} & 情况特别严重，因为该法律在许多方面是其他法律的基础。 \\
		  & (The situation is particularly serious, because in many ways this law is the basis of other laws.) \\
		  \textbf{Ours} & 情况\textcolor{blue}{要}$_1$特别\textcolor{blue}{重大}$_2$的是因为这种法律在\textcolor{blue}{许多程度}$_3$上是\textcolor{blue}{构建}$_4$其他法律的基础\textcolor{blue}{所在}$_5$。 \\
		   & (The situation \textcolor{blue}{needs}$_1$ to be particularly \textcolor{blue}{significant}$_2$ because \textcolor{blue}{to many extent}$_3$ this law is the basis for \textcolor{blue}{where}$_5$ the \textcolor{blue}{construction}$_4$ of other laws.) \\
		   \hline
		   \textbf{Source}  & 总之，她们的生活质量非常低。 \\
		   & (In short, their quality of life is very poor.) \\
		   \textbf{BT} & 总之她们的生活质量非常\textcolor{blue}{不好}$_1$。\\
		   & (In short their quality of life is very \textcolor{blue}{bad}$_1$.) \\
		   \textbf{Rule} & 总，之她\textcolor{blue}{闷}$_1$的生活\textcolor{blue}{效率}$_2$非常低。 \\
		   & (In, short her \textcolor{blue}{boring}$_1$ life's \textcolor{blue}{efficiency}$_2$ is very poor.) \\
		   \textbf{RT} & \textcolor{blue}{简言}$_1$之，他们的生活质量\textcolor{blue}{很}$_2$低。 \\
		   & (\textcolor{blue}{In short}, their quality of life is \textcolor{blue}{very} poor.) \\
		   \textbf{Ours} & \textcolor{blue}{可以说}$_1$，她们\textcolor{blue}{生存}$_2$\textcolor{blue}{素质}$_3$非常\textcolor{blue}{弱}$_4$。\\
		   & (\textcolor{blue}{So to speak}$_1$, their \textcolor{blue}{survival}$_2$ \textcolor{blue}{quality}$_3$ is very \textcolor{blue}{weak}$_4$.) \\
		   \hline
		\end{tabular}
		
		\caption{\label{example} Examples of synthetic erroneous sentences. The rewritten tokens are highlighted in \textcolor{blue}{blue}.}
	\end{table}
\end{CJK*}

To give a qualitative analysis of generated erroneous sentences, we present examples of our approach and existing synthetic methods in Table~\ref{example}. We can see that back-translation tends to generate similar modifications such as token deletion and simple paraphrasing. The rule-based corruption approach is hard to simulate human writing since it directly swaps adjacent tokens and performs word replacement without consideration of the sentence context. Round-trip translation generates error-free sentences. In contrast, our approach generates less fluent sentences by paraphrasing the corrupted contents and maintaining the meaning preserved in the corresponding English sentences. 


%% file: sections/related.tex
\section{Related Work}

Pretraining a seq2seq model on synthetic data and then fine-tuning on annotated error-corrected sentence pairs is the common practice for GEC. Available datasets in non-English languages such as German~\citep{boyd2018using}, Russian~\citep{rozovskaya2019grammar}, Czech~\citep{naplava2019grammatical}, and Spanish~\citep{davidson2020developing} only contain less than 50000 pairs, which results in a high requirement on the quality of synthetic data construction. 

Back-translation (BT) is the reverse of GEC, which takes corrected sentences as input and error sentences as output. It is popular and effective for English GEC \citep{ge2018fluency,xie2018noising,kiyono2019empirical,wan2020improving,stahlberg2021synthetic} but difficult to adapt to these low-resource scenarios, since it is hard to learn diverse error-corrected patterns with less annotated sentence pairs. Round-trip translation (e.g., translating German to English then back to German) \citep{lichtarge2019corpora} has been blamed for errors that it introduced are relatively clean and achieves inferior performance in the non-English language as shown in our experiment. 

The most effective construction method on non-English languages is rule-based corruption~\citep{naplava2019grammatical,grundkiewicz2019minimally}. Most of them rely on word substitutions with ASpell or language-specific confusion sets. It requires well-designed rules to simulate diverse linguistic phenomena across different languages. \cite{rothe2021simple} only performs language-agnostic operations without any confusion set to construct corrupted sentences but achieves inferior performance with moderate model size. Wikipedia edits extracted from the revision history of each page is another useful GEC pretraining resource \citep{boyd2018using,flachs2021data}. Most studies for Chinese GEC focus on model architecture \citep{fu2018youdao,zhou2018chinese,ren2018sequence} or training strategy \citep{zhao2020maskgec}, which are orthogonal with our approach.


The most similar approach to ours is \cite{zhou2020improving} which trains two autoregressive translation models with poor and good qualities, respectively. With the same sentence in the source language, they regard two translations of two models as error-corrected sentence pairs. In comparison, our approach directly utilizes the non-autoregressive translation ability of the PXLM without training translators additionally, which is easier to adapt to new languages. 
Utilizing a pretrained language model to propose candidate words for replacement and insertion has also been applied to lexical substitution \citep{zhou2019bert}, text generation \citep{NEURIPS2020_7a677bb4}, etc. By contrast, we adopt bilingual constraint to avoid inconsistency of candidate words with the original meaning, which plays an important role for data construction of GEC as shown in our experiment.


How to leverage pretrained language models in GEC seq2seq models has been extensively explored \citep{katsumata2020stronger,wang2020chinese,kaneko2020encoder}. In this paper, we initialize the model with DeltaLM \citep{ma2021deltalm}, which adjusts the InfoXLM-initialised encoder-decoder model to generation mode by self-supervised pretraining. 

%% file: sections/conclusion.tex
\section{Conclusion and Future Work}
 We propose a unified and generic strategy for training GEC systems in non-English languages given a PXLM and the parallel translation data. Our approach obtains state-of-the-art results on the NLPCC 2018 Task 2 dataset (Chinese) and competitive results on German and Russian benchmarks. The synthetic sentence pairs also complement rule-based corruption to yield further improvement. Compared with a regular translator, NAT by the PXLM can control the degree of overlap between the generated sentence and the original sentence. The bilingual constraint also ensures that the sampled tokens will not deviate from the original meaning, which plays an important role in our strategy.
 
 
We plan to investigate whether utilizing back-translated English sentences rather than gold English sentences leads to similar performance, which could get rid of quantitative restriction by the size of translation corpus to generate the unlimited number of error-corrected sentence pairs. 

%% file: sections/appendix.tex
\appendix
\begin{table*}[h]
    \centering
    \begin{tabular}{l|ccccc|cccccc}
    \multirow{2}{*}{\textbf{Language}} & \multicolumn{5}{c|}{\textbf{Token-level Operations}} & \multicolumn{6}{c}{\textbf{Post Edit}} \\
     & $p_{noise}$ & mask & insert & delete & swap & $p_{noise}$ & substitute & insert & delete & swap & recase \\
    \hline
    Chinese & 0.5 & 0.7 & 0.1 & 0.1 & 0.1 & 0.05 & 0.3 & 0.2 & 0.3 & 0.2 & 0 \\
    German & 0.3 & 0.65 & 0.15 & 0.15 & 0.05 & 0.02 & 0.25 & 0.25 & 0.2 & 0.2 & 0.1 \\
    Russian & 0.15 & 0.65 & 0.15 & 0.15 & 0.05 & 0.02 & 0.25 & 0.25 & 0.2 & 0.2 & 0.1 \\
    \hline
    \end{tabular}
    \caption{Parameters for NAT-based data construction and post edit.}
    \label{gen}
\end{table*}

\begin{table}[h]
\centering
\small
\begin{tabular} {lr} 
\hline
Configurations	         &	Values				\\
\hline
\multicolumn{2}{c}{\bf Pretraining} \\
\hline
Model Architecture 	    & Transformer (base) \\
Devices                 & 8 Nvidia V100 GPU      \\
Max tokens per GPU		& 5120					\\
Update Frequency        & 8                     \\
Optimizer 				& Adam 					\\
						& ($\beta_1$=0.9, $\beta_2$=0.98, $\epsilon$=$1\times10^{-8}$)	\\
Learning rate 			& $7\times10^{-4}$			\\
Learning rate scheduler & polynomial decay \\
Warmup                  & 8000 \\
weight decay            & 0.0 \\
Loss Function 			& label smoothed cross entropy \\
						& (label-smoothing=0.1) \\
Dropout					& 0.3	\\
\hline
\multicolumn{2}{c}{\bf Fine-tuning} \\
\hline
Devices                 & 4 Nvidia V100 GPU      \\
Update Frequency        & [2, 4]                     \\
Learning rate 			& [$5\times10^{-4}$, $7\times10^{-4}$]			\\
Warmup                  & 4000 \\

\hline
\end{tabular}
\caption{Hyper-parameters values of pretraining and fine-tuning on NLPCC 2018 Task 2 (Chinese).\label{chinese}} 
\end{table}

\begin{table}[h]
\centering
\small
\begin{tabular} {lr} 
\hline
Configurations	         &	Values				\\
\hline
\multicolumn{2}{c}{\bf Pretraining} \\
\hline
Model Architecture 	    & Transformer (base) \\
Devices                 & 8 Nvidia V100 GPU      \\
Max tokens per GPU		& 5120					\\
Update Frequency        & 8                     \\
Optimizer 				& Adam 					\\
						& ($\beta_1$=0.9, $\beta_2$=0.98, $\epsilon$=$1\times10^{-8}$)	\\
Learning rate 			& [$5\times10^{-4}$, $7\times10^{-4}$]			\\
Learning rate scheduler & polynomial decay \\
Warmup                  & 8000 \\
weight decay            & 0.0 \\
Loss Function 			& label smoothed cross entropy \\
						& (label-smoothing=0.1) \\
Dropout					& [0.1, 0.2]	\\
Source Dropout          & 0.2   \\
Target Dropout          & 0.1   \\
Edit-weighted MLE       & 3     \\
\hline
\multicolumn{2}{c}{\bf Fine-tuning} \\
\hline
Devices                 & 1 Nvidia V100 GPU      \\
Update Frequency        & [2, 4]                     \\
Learning rate 			& [$3\times10^{-4}$, $5\times10^{-4}$, $7\times10^{-4}$]			\\
Dropout					& [0.1, 0.2, 0.3]	\\
Warmup                  & 2000 \\

\hline
\end{tabular}
\caption{Hyper-parameters values of pretraining and fine-tuning on Falko-Merlin (German) and RULEC-GEC (Russian).\label{de_ru}} 
\end{table}

\section{Hyper-parameters}

The parameters for NAT-based data construction and post edit are presented in Table \ref{gen}. The hyper-parameters of training the Transformer on NLPCC 2018 Task 2 (Chinese) are listed in Table \ref{chinese}. The hyper-parameters for German and Russian are shown in Table \ref{de_ru}. For Russian, we use rule-based corruption without any language-specific operation and confusion set with 50\% probability to assist in synthetic data construction.